\documentclass{article} 
\usepackage{iclr2027_conference,times}

\usepackage{amsmath,amsfonts,bm}

\def\eqref#1{equation~\ref{#1}}

\def\1{\bm{1}}

\DeclareMathAlphabet{\mathsfit}{\encodingdefault}{\sfdefault}{m}{sl}
\SetMathAlphabet{\mathsfit}{bold}{\encodingdefault}{\sfdefault}{bx}{n}

\usepackage{url}
\usepackage{graphicx}
\usepackage{float}
\usepackage{needspace}
\usepackage{xspace}
\usepackage{adjustbox}
\usepackage{wrapfig}
\AddToHook{env/figure/begin}{\setlength{\abovecaptionskip}{8pt}}
\AddToHook{env/figure*/begin}{\setlength{\abovecaptionskip}{8pt}}
\AddToHook{env/table/begin}{\setlength{\belowcaptionskip}{6pt}}
\AddToHook{env/table*/begin}{\setlength{\belowcaptionskip}{6pt}}
\usepackage{lipsum}
\usepackage{multirow}
\usepackage{makecell}
\usepackage{pifont}
\usepackage{algorithmic}
\usepackage{listings}

\newcommand{\cmark}{\ding{51}}
\newcommand{\xmark}{\ding{55}}
\DeclareRobustCommand{\circlednum}[1]{\ding{\numexpr171+#1\relax}}

\newcommand{\name}[0]{AKV\xspace}
\newcommand{\xhdr}[1]{{\noindent\bfseries #1}.}

\newcommand{\rom}[1]{\textup{\uppercase\expandafter{\romannumeral#1}}}

\newcounter{observation}

\newcounter{design}

\usepackage[most]{tcolorbox}
\usepackage{colortbl}
\usepackage{pifont}
\usepackage{makecell}

\tcbuselibrary{skins, breakable}   
\usepackage{tabularx, booktabs}    

\definecolor{darkred}{rgb}{0.5,0,0}
\definecolor{darkgreen}{rgb}{0,0.5,0}
\definecolor{tahgray}{RGB}{242,242,242}

\newtcolorbox[
  auto counter,
  list inside=examplelist,
]{greenbox}[2][]{
  enhanced,
  title={Example~\thetcbcounter. \textbf{#1}},
  colback=green!5!white,
  colframe=green!50!black,
  colbacktitle=green!60!black,
  coltitle=white,
  #2
}

\newtcolorbox[
  auto counter,
  list inside=examplelist
]{bluebox}[2][]{
  enhanced,
  title={Text~\thetcbcounter. \textbf{#1}},
  colback=cyan!5!white,  
  colframe=cyan!50!black,  
  colbacktitle=cyan!75!black,  
  coltitle=white,
  #2
}

\newcommand{\M}{\mathrm{Mem}}

\usepackage{hyperref}
\usepackage{amssymb}

\title{Cache Engineering for Agents}

\author{\textbf{Tianyu Fu}\thanks{Equal contribution.\quad $^{\dagger}$Corresponding author: Yu Wang (yu-wang@tsinghua.edu.cn).} \ $^{1}$,
    \textbf{Ruoxi Wang}$^{*1}$,
    \textbf{Zihan Min}$^{*1}$,
    \textbf{Zhongqi Ren}$^{1}$,
    \textbf{Zhihao Li}$^{1}$,
    \\
    \textbf{Tengxuan Liu}$^{1}$,
    \textbf{Shengxu Mu}$^{1}$,
    \textbf{Hanling Zhang}$^{2}$,
    \textbf{Yu Wang}$^{\dagger1}$
    \\
    \\
    $^{1}$Tsinghua University \quad
    $^{2}$The Chinese University of Hong Kong
}

\preprintcopy
\renewcommand{\iclrruler}[1]{} 
\begin{document}

\maketitle

\begin{abstract}
\label{sec:abstract}
Large Language Model (LLM) agents can handle increasingly complex tasks, navigating long trajectories of observations and actions.
As these trajectories grow, key-value (KV) cache I/O and storage become the dominant cost.
Trajectories can also exceed the LLM's maximum context length, requiring control over active cache growth.
Prior work uses \textit{context engineering} to reduce cache size by trimming or summarizing history.
However, modifying the trajectory prefix invalidates subsequent cached states and triggers costly re-prefilling of retained context.
Task accuracy can also suffer when too little context is retained to limit re-prefill costs.
We propose \textit{cache engineering} as a new scheme. 
It allows agents to directly control their active KV cache without modifying the trajectory or re-prefilling retained context.
Specifically, agents can proactively use $\operatorname{drop}$ to exclude selected cache from future attention and $\operatorname{repos}$ to adjust retained keys' position encodings.
We formalize these cache operations and identify history dependence as a serving challenge: identical token sequences can yield different cache states under different cache operation sequences, making prefix matching insufficient for correct cache reuse.
Thus, we co-design the \name serving engine. 
It provides a stateless interface compatible with existing serving APIs, while internally managing stateful cache reuse, eviction, and restoration.
Experiments show that a tool-response rolling-drop cache policy improves average benchmark scores by 1.7 points and reduces average cost by 27.7\%, compared with conventional threshold-triggered summarization.
With \name, it achieves up to $1.2\times$ the output throughput and a $1.5\times$ speedup in P95 request latency.
Recursive self-improvement of cache policies further reduces cost by 7.1\% over expert design.
Our code is at \url{https://github.com/thu-nics/AKV}.
\end{abstract}





\section{Introduction}
\label{sec:intro}

Large Language Model (LLM) agents are capable of increasingly long-horizon tasks.
Agents are generally composed of an LLM for reasoning and tool calling, and a harness that orchestrates tool execution and context. 
Through repeated reasoning, tool use, and environment interactions, they can carry out complex work spanning hundreds of rounds~\citep{ye2025agentfold, liu2025contexttool}.

As agent trajectories grow longer, Key-Value (KV) cache growth poses challenges to both inference cost and task accuracy.
In terms of cost, retaining and repeatedly accessing accumulated KV caches incur substantial storage and I/O overhead~\citep{wu2026dualpath, anthropic2026fable}. As shown in Figure~\ref{fig:bcp-cost-breakdown}, KV cache reads take up 82.4\% of the total API cost.
In terms of task accuracy, agents can suffer severe degradation if either the number of attended KV states or their positional range exceeds what LLM handles reliably~\citep{hsieh2024ruler, chen2026iterresearch, zhou2025mem1}.
Therefore, it requires delicate control over the active KV cache for efficient and reliable agent execution.

Prior work reduces the growing cache via \textit{context engineering}, typically by trimming or summarizing historical messages.
While older messages are often less useful, editing them changes the trajectory prefix, invalidates subsequent KV states, and triggers re-prefill.
The earlier the edit, the more retained trajectory suffix that must be re-prefilled.
Common harnesses therefore defer reductions until an active context length threshold is reached, then trim or summarize most of the trajectory and retain only a few recent rounds to keep re-prefill short~\citep{2026pi,openai2026codex}.
Such aggressive reductions risk losing useful information from intermediate rounds, exposing a dilemma between re-prefill cost and task accuracy.

\begin{figure}[t]
    \centering
    \includegraphics[width=\linewidth]{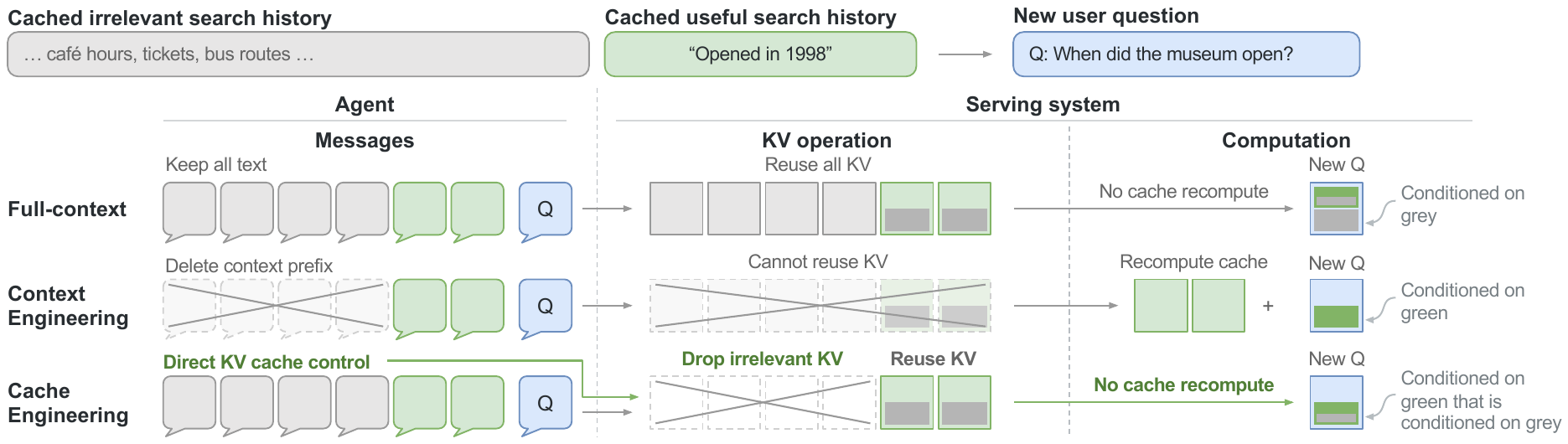}
    \caption{Full-context inference, context engineering, and cache engineering.
    Full-context inference reuses all historical KV.
    Context engineering removes irrelevant text and recomputes the retained history.
    Cache engineering keeps the text history but drops selected KV entries, reusing retained states without recomputation.
    Nested color blocks illustrate how retained KV can still carry information from dropped history.}
    \label{fig:scheme}
\end{figure}

In this paper, we introduce \textit{cache engineering}, a new scheme for agents to directly control their active KV cache.
As shown in Figure~\ref{fig:scheme}, instead of modifying the text trajectory, it excludes only selected KV entries from future attention, and reuses all retained KV states without re-prefill.
This allows agents to incrementally drop older KV entries while retaining more suffix KV within a given cost budget.
To support this control, alongside the usual $\operatorname{append}$ operation that adds KV states for new tokens, we introduce two additional cache operations. $\operatorname{Drop}$ removes selected KV entries from future attention, and $\operatorname{repos}$ adjusts the positions encoded in retained keys.
Throughout execution, the LLM and harness choose when and how to apply these operations, repeatedly interleaving them with $\operatorname{append}$ to shape the active cache.
This programmable control also enables agents to search, evaluate, and refine cache engineering strategies through recursive self-improvement (RSI).

However, the history-dependent nature of cache engineering challenges current LLM serving systems.
With standard KV caching, the token prefix fully determines the cached states~\citep{kwon2023efficient, zheng2024sglang}.
In contrast, the same token prefix and retained positions can result in different cache states under different cache operation histories.
As shown in Figure~\ref{fig:drop}, trajectories \circlednum{1} and \circlednum{2} both retain KV entries for B and C, but C's KV is computed with attention to A only in \circlednum{1}, due to the different ordering of drop and append operations.
The serving system must therefore account for both token and cache operation histories to correctly track or reuse cache states.

Thus, we co-design \name, a user-friendly and efficient serving system for cache engineering.
It provides a \textit{stateless} interface compatible with existing API-based serving, while handling the complexity of \textit{stateful} cache management internally.
Agents simply submit messages as usual, accompanied by cache operation instructions, without tracking which KV states remain resident on the server.
The system uses an extended radix tree for correct reuse and timely eviction of history-dependent cache states, while segment-batch attention efficiently reconstructs missing states in a single prefill pass.
\name turns the theoretical efficiency potential of cache engineering into lower wall-clock latency and higher throughput for real-world agent workloads.

\begin{itemize}
    \item \textbf{Cache engineering formulation.}
    We formalize cache engineering through explicit cache operations and their trajectories, providing a clear and deterministic description of history-dependent cache states.
    This formulation also provides a programmable interface for agents to define and refine their own cache engineering strategies.
    \item \textbf{Serving system for cache engineering.} 
    We propose the first serving system that allows agents to directly control their caches, which were previously managed solely by the system. 
    We co-design a stateless interface, an extended radix tree, and segment-batch attention to support efficient cache reuse, eviction, and restoration.
    \item \textbf{Improving efficiency and benchmark performance.} 
    Across three LLMs and four agent benchmarks, a simple tool-response rolling-drop cache policy reduces average cost by 27.7\% and improves average benchmark scores by 1.7 points, compared with conventional threshold-triggered summarization. 
    With \name server, this translates to $1.18\times$ the output throughput and a $1.51\times$ speedup in P95 request latency.
    An RSI-based policy additionally reduces cost by 7.1\% on held-out tasks.
\end{itemize}

\section{Related Work}
\label{sec:related_work}

\xhdr{Agent context engineering}
At the text level, context engineering summarizes or trims interaction histories to reduce cost and enable agents to work beyond the LLM's context window.
Existing methods trigger summarization through heuristics~\citep{wu2025resum} or agent decisions~\citep{ye2025agentfold, sun2025scaling}.
Further work improves summary quality through reinforcement learning~\citep{lu2025scaling} or by optimizing compression guidelines~\citep{kang2025acon}.
To avoid the cost and complexity of summarization, trimming methods~\citep{lindenbauer2025complexity,zhang2026masking} directly remove less relevant context.
However, editing earlier context requires re-prefilling the retained suffix, creating a dilemma between re-prefill cost and information retention.

\xhdr{KV cache compression, eviction, and reuse}
At the cache level, prior work reduces cache size or approximately reuses existing states to lower storage, access, and computation costs.
\textit{Compression} methods use low-rank representations~\citep{liu2024deepseek, wang2024cskv}, quantization~\citep{liu2024kivi,liu2026pm}, or aggregation~\citep{fu2025framefusion,wang2024model} to represent cached entries more compactly.
\textit{Eviction} methods typically rely on low-level signals such as token positions~\citep{xiao2024efficient} or attention-based importance~\citep{zhang2023h2o, li2024snapkv, fu2025moa, yang2026indexmem} to remove less useful cache entries.
\textit{Reuse} methods adapt cached states to different positions or context prefixes through position encoding adjustments~\citep{gao2024cost, ma2026irminsul,ordonez2026minipic} or partial recomputation~\citep{wang2025mepic}.
Our work focuses on eviction and reuse, enabling high-level eviction decisions by agents and harnesses, while providing a practical serving engine for efficient cross-request reuse and isolation.
We discuss concurrent agent-oriented eviction work in Appendix~\ref{sec:concurrent_works}.

\xhdr{KV cache management in serving systems}
Beyond algorithmic designs, previous works also explore system optimizations for efficient cache storage, sharing, and request scheduling.
Existing systems use prefix caching to share KV states across requests with common prefixes while preserving equivalence to full prefill~\citep{zheng2024sglang, ye2024chunkattention}.
For agent workloads, recent systems batch context edits to reduce cache invalidation~\citep{xu2026tokenpilot} or overlap re-prefill with ongoing execution to hide its latency~\citep{pan2026smoothagent}.
For cache engineering, it additionally requires tracking cache operation histories, since identical token prefixes can produce different states under different operations.


\section{Cache Engineering}
\label{sec:formulation}

Cache engineering lets agents directly control their active KV cache without rewriting the trajectory or re-prefilling retained context.
We introduce basic concepts in Section~\ref{sec:preliminaries} and design motivations in Section~\ref{sec:motivation}, then formalize cache operations and their trajectories in Sections~\ref{sec:cache_operation} and~\ref{sec:cache_trajectory}.

\subsection{Preliminaries}
\label{sec:preliminaries}

An autoregressive LLM predicts the next token conditioned on the preceding token sequence $x_{1:t}$:
\begin{equation}
    p_\theta(x_{t+1}\mid x_{1:t}).
    \label{eq:token-ntp}
\end{equation}
Modern LLM serving uses KV caching to avoid recomputing earlier tokens' states.
The next-token prediction is then conditioned on the current token $x_t$ and the preceding tokens' cache:
\begin{equation}
    p_\theta(x_{t+1}\mid x_t,\mathcal{C}_{1:t-1}),
    \label{eq:cached-ntp}
\end{equation}
where $\mathcal{C}_{1:t-1}$ denotes the KV states computed from tokens $x_{1:t-1}$.



An agent's \emph{trajectory} is the ordered text history used for the following model computation, including its prompt, assistant messages, and tool responses.
Formally, the trajectory is a \emph{logical token sequence} $x_{1:t}$, and $t$ is its \emph{logical token length}.

In cache engineering, the cache size need not equal the logical token length.
We therefore define the \emph{active cache} $\mathcal{C}$ as the KV cache states available to subsequent attention.
The corresponding token sequence is the \emph{active context}, whose length equals the number of token entries in $\mathcal{C}$.
Note that \emph{active} refers to participation in attention at the algorithmic level, not physical cache residency in the serving engine.
A cache entry's \emph{position encoding index} can also be adjusted independently of its token's \emph{logical index} in the trajectory.

\textit{Context engineering} modifies the logical token sequence through trimming or summarization.
In contrast, \textit{cache engineering} directly modifies the active cache through cache operations, while keeping the logical token sequence unchanged.

\subsection{Design Motivation}
\label{sec:motivation}

The design of cache engineering stems from our thinking about three key questions: why direct cache modification is needed, which parts of the cache to retain, and how to manage its encoding positions. We discuss the motivations and empirical evidence guiding our design decisions below.

\begin{figure}[t]
    \centering
    \begin{minipage}[t]{0.32\linewidth}
    \vspace{0pt}
    \centering
    \includegraphics[width=\linewidth,height=0.64\linewidth,keepaspectratio]{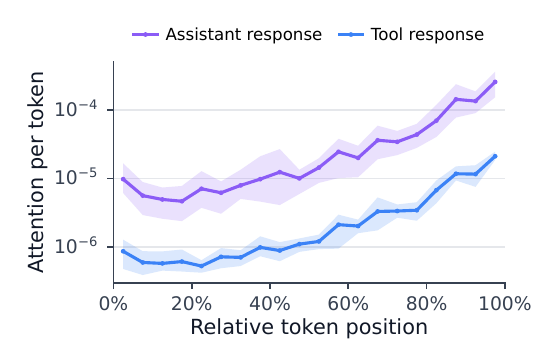}
    \caption{Attention over history.
    Per-token attention to assistant and tool responses, averaged over 10 BCP requests.}
    \label{fig:bcp-attention-distance-curve}
\end{minipage}\hfill
  \begin{minipage}[t]{0.32\linewidth}
    \vspace{0pt}
    \centering
    \includegraphics[width=\linewidth,height=0.64\linewidth,keepaspectratio]{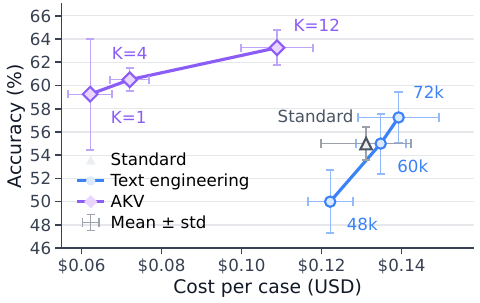}
    \caption{GPT-OSS-120B accuracy and cost on 100 BCP tasks.
    Labels indicate summary thresholds (tokens) or \name's retained tool rounds.}
    \label{fig:pareto-frontier}
  \end{minipage}\hfill
  \begin{minipage}[t]{0.32\linewidth}
    \vspace{0pt}
    \centering
    \includegraphics[width=\linewidth,height=0.64\linewidth,keepaspectratio]{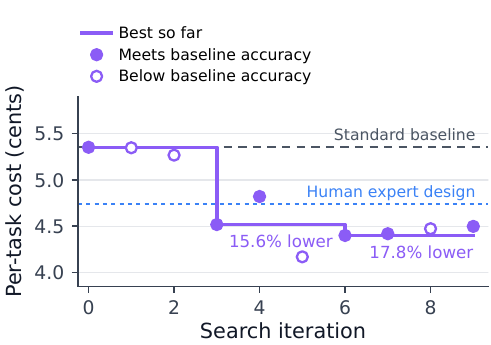}
    \caption{Recursive self-improvement on cache engineering policy, evaluated on the BCP benchmark with MiniMax-M2.7 LLM.}
    \label{fig:minimax-self-evolution} 
  \end{minipage}
\end{figure}

\xhdr{Why modify the cache directly}
Cache engineering alleviates the dilemma between re-prefill cost and task accuracy in context engineering.
We evaluate GPT-OSS-120B on the first 100 BCP tasks, with four trials per task and at most 100 agent rounds.
Trim-based context engineering (retaining only the latest 12 rounds of tool responses) costs $1.85\times$ as much as the full-context baseline.
Summary-based context engineering (48k threshold) reduces cost by about 7\% but lowers accuracy by 5 percentage points (Figure~\ref{fig:pareto-frontier}).
In contrast, directly dropping older tool-response cache can reuse the remaining cache, simultaneously improving accuracy and reducing cost (Figure~\ref{fig:pareto-frontier}).
The retained cache may also preserve information from dropped history used in its computation~\citep{xiao2024efficient,fu2026c2c,zou2025latent}, as shown in Figure~\ref{fig:scheme}.

\xhdr{Which parts of the cache to retain}
Cache selection should reflect message roles, recency, and task needs, beyond token-level importance heuristics.
Figure~\ref{fig:bcp-attention-distance-curve} shows that older tokens receive less attention than recent ones, and tool responses receive less attention than assistant messages at comparable relative positions.
We therefore use tool-response rolling drop as our primary example, retaining cache for all reasoning and recent tool responses while dropping older tool-response cache.
We later explore alternative policies, including RSI-based cache engineering.

\xhdr{How to manage position encodings}
By default, position encoding indices grow with the logical trajectory, even as the cache is reduced.
Dropping cache neither changes retained position encodings nor resets the next token's position index, so indices can reach the model's limit even with a small cache.
We decouple position encoding indices from logical indices and cache size, using a separate cache operation to reposition cached keys with RoPE rotations~\citep{su2023roformer} without re-prefill.

\subsection{Cache Operations}
\label{sec:cache_operation}

\begin{figure}[t]
    \centering
    \includegraphics[width=\linewidth]{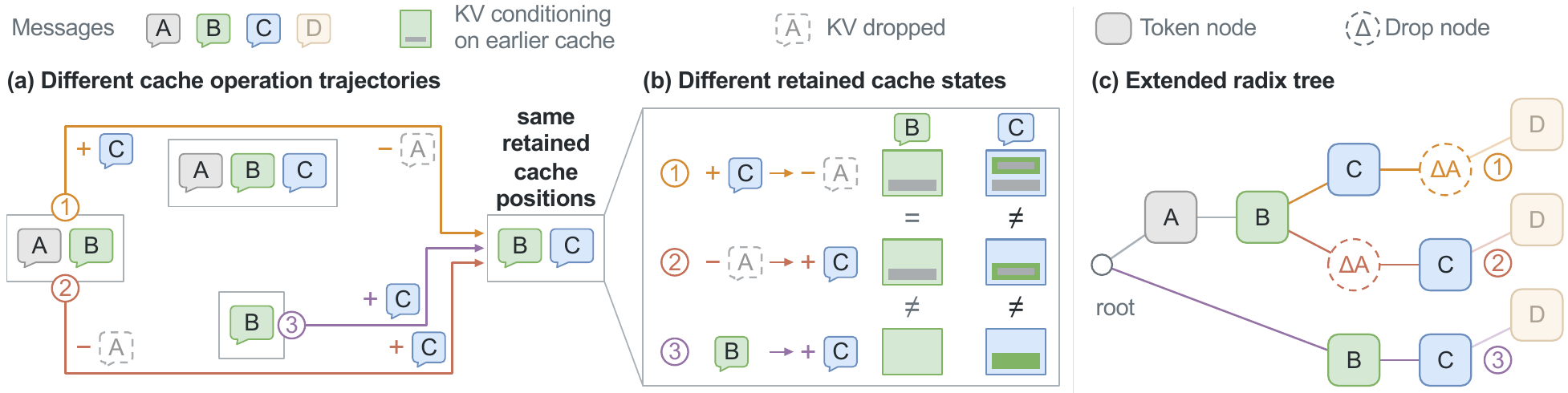}
    \caption{Cache operation trajectories and history-dependent KV states.
    (a) \circlednum{1} appends C to A and B, then drops A's KV;
    \circlednum{2} drops A's KV before appending C;
    \circlednum{3} starts from B alone and appends C.
    All three retain B and C.
    (b) B's KV is identical in \circlednum{1} and \circlednum{2}, both conditioned on A, but differs in \circlednum{3}, where B has no prefix.
    C's KV differs in all three: \circlednum{1} is computed with A and B; \circlednum{2} and \circlednum{3} cannot attend to A but use different B caches.
    (c) The extended radix tree shares the A--B prefix of \circlednum{1} and \circlednum{2}, then branches according to $\operatorname{drop}$ timing, distinguishing the cached states used to generate D.}
    \label{fig:drop}
\end{figure}

We define three basic cache operations for cache engineering: $\operatorname{append}$, $\operatorname{drop}$, and $\operatorname{repos}$, which directly manipulate cache without rewriting the existing logical token sequence.

\xhdr{$\operatorname{Append}$}
Operation $\operatorname{append}$ is the standard cache update in LLM inference.
It processes a new token $x_{t+1}$ conditioned on the active cache and appends its cache entry $C_{t+1}$:
\begin{equation}
    \operatorname{append}(x_{t+1}):\qquad
    \mathcal{C}\leftarrow[\mathcal{C};\,C_{t+1}].
    \label{eq:append-operation}
\end{equation}
We represent $\mathcal{C}$ as a sequence of KV entries ordered by their logical indices, where $[\cdot;\cdot]$ denotes concatenation.
The token may come from the prompt, a tool response, or model generation.
Position encoding indices start at zero and advance by one after each append; $\operatorname{repos}$ can reset the next index.

\xhdr{$\operatorname{Drop}$}
Operation $\operatorname{drop}$ removes selected KV states from subsequent attention, while leaving the retained states, their position encodings, and the logical token sequence unchanged (Figure~\ref{fig:scheme}):
\begin{equation}
    \operatorname{drop}(P):\qquad
    \mathcal{C}\leftarrow\mathcal{C}\setminus\{C_i:i\in P\},
    \label{eq:drop-operation}
\end{equation}
where $P$ is the set of active logical token indices selected for dropping, and $\setminus$ denotes cache removal, preserving the order of retained entries.
Note that $\operatorname{drop}$ does not change the position encoding indices used by subsequent $\operatorname{append}$ operations.


\xhdr{$\operatorname{Repos}$}
Operation $\operatorname{repos}$ updates cached keys' position encodings in logical token order, and resets the position index for subsequent $\operatorname{append}$ operations.
With $r$ retained cache entries, their position indices become $0,\ldots,r-1$, and the next appended token uses index $r$.
For rotary position embeddings (RoPE)~\citep{su2023roformer}, we move a cached key $K$ from position index $p$ to $q$ by
\begin{equation}
    K\leftarrow R(q)R(p)^{-1}K,
    \label{eq:repos-operation}
\end{equation}
where $R(p)$ is the rotation matrix at position $p$.
Cached values remain unchanged.


These operations define the cache visible to attention at the algorithmic level.
The serving system handles physical storage, eviction, and reuse of cache separately in Section~\ref{sec:serving}.

\subsection{Cache Operation Sequence}
\label{sec:cache_trajectory}

To provide a clear and deterministic description of cache operations, we use a \emph{cache operation sequence}:
\begin{equation}
    o_{1:m}=(o_1,\ldots,o_m),
    \qquad o_j\in\{\operatorname{append}(x),\operatorname{drop}(P),\operatorname{repos}\}.
    \label{eq:operation-sequence}
\end{equation}
Here, $m$ counts operations, $x$ is a token, and $P$ contains the active logical token indices to drop.
For a fixed LLM, this sequence determines the current cache states, starting from an empty cache.

Cache operation sequences are necessary because cache states are history-dependent: tokens and retained positions alone do not determine next-token probabilities.
For example, in Figure~\ref{fig:drop}, all three trajectories retain $B$ and $C$, but $A$'s cache is available when computing both $B$ and $C$ in \circlednum{1}, only $B$ in \circlednum{2}, and neither in \circlednum{3}. Thus, $C$'s cache states are different in all three trajectories.
Similarly, $\operatorname{repos}$ changes the position encodings used by later $\operatorname{append}$ operations, as shown in Figure~\ref{fig:repos}.
Correct cache reuse and restoration therefore require both token and cache operation histories.


\section{\name Serving System}
\label{sec:serving}

\name follows the principle of \textit{stateless interface, stateful serving}.
Agents control the cache used in subsequent attention by specifying cache operations (Section~\ref{sec:interface}), while the server handles physical cache reuse, eviction, and restoration internally (Section~\ref{sec:cache-management}).

\subsection{Serving Interface}
\label{sec:interface}

\begin{figure}[tb]
    \centering
    \begin{minipage}[t]{0.68\linewidth}
    \vspace{0pt}
    \centering
    \begin{minipage}[c][1.38in][c]{\linewidth}
    \centering
    {\footnotesize\fontencoding{T1}\selectfont
    \renewcommand{\arraystretch}{1.0}
    \begin{tabular}{@{}l@{\enspace}l@{}}
    \{\enspace\textquotedbl{}model\textquotedbl{}: \textquotedbl{}gpt-oss-120b\textquotedbl{}, &  \\
    \hspace*{1em}\textbf{\textquotedbl{}messages\textquotedbl{}}: [ &  \\
    \hspace*{2em}\{\textquotedbl{}role\textquotedbl{}: \textquotedbl{}user\textquotedbl{}, \textquotedbl{}content\textquotedbl{}: \textquotedbl{}Search ICLR\textquotedbl{}\}, & $\triangleright\ \operatorname{append}(m_0)$ \\
    \hspace*{2em}\{\textquotedbl{}role\textquotedbl{}: \textquotedbl{}assistant\textquotedbl{}, \textquotedbl{}content\textquotedbl{}: \textquotedbl{}Searching\textquotedbl{}\}, & $\triangleright\ \operatorname{append}(m_1)$ \\
    \hspace*{2em}\{\textquotedbl{}role\textquotedbl{}: \textquotedbl{}tool\textquotedbl{}, \textquotedbl{}content\textquotedbl{}: \textquotedbl{}ICLR is...\textquotedbl{}\} & $\triangleright\ \operatorname{append}(m_2)$ \\
    \hspace*{1em}], \textbf{\textquotedbl{}cache\textquotedbl{}}: \{ &  \\
    \hspace*{2em}\textbf{\textquotedbl{}drop\textquotedbl{}}: \{\textquotedbl{}1\textquotedbl{}: [0], & $\triangleright\ \operatorname{drop}(m_0)$ \\
    \hspace*{2em}\phantom{\textbf{\textquotedbl{}drop\textquotedbl{}}: \{}\textquotedbl{}2\textquotedbl{}: [1]\}, & $\triangleright\ \operatorname{drop}(m_1)$ \\
    \hspace*{2em}\textbf{\textquotedbl{}reposition\textquotedbl{}}: [2] & $\triangleright\ \operatorname{repos}$ \\
    \}\,\} & \\
    \end{tabular}
    }
    \end{minipage}
    \caption{Illustrative \name request. Right-hand annotations show cache operations; $m_i$ denotes message $i$.}
    \label{fig:cache-request}
    \end{minipage}\hfill
    \begin{minipage}[t]{0.30\linewidth}
    \vspace{0pt}
    \centering
    \begin{minipage}[c][1.38in][c]{\linewidth}
    \centering
    \includegraphics[width=\linewidth,height=1.38in,keepaspectratio]{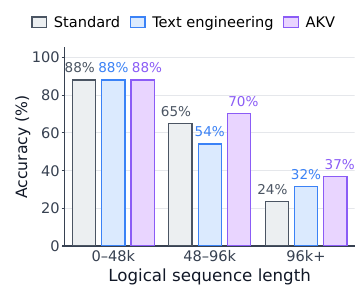}
    \end{minipage}
        \caption{BCP accuracy by length with GPT-OSS-120B.}
        \label{fig:bcp-correct-cost-by-length}
    \end{minipage}
\end{figure}

The serving interface must keep each request's output distribution independent of physical cache residency and other requests' cache operations.
For example, $\operatorname{drop}$ must affect only the issuing request's attention, leaving other requests' output distributions unchanged even when they share the same prefix cache.
Each request must also fully describe the cache states it uses, without relying on physical cache retained from previous turns.
This allows the server to fully restore missing cache states if memory pressure causes eviction between turns.

To meet these requirements, \name extends the standard LLM serving API to specify both token and cache operation histories.
As shown in Figure~\ref{fig:cache-request}, the existing \texttt{messages} field provides the full logical token sequence, while the optional \texttt{cache} field specifies $\operatorname{drop}$ and $\operatorname{repos}$ operation sequence using an \textit{operate-after} grammar.
Note that all entries in \texttt{messages} implicitly specify $\operatorname{append}$ operations.
Each key in \texttt{drop} identifies the message index after which to apply the operation; its value lists the message indices whose cache is dropped.
The \texttt{reposition} field lists the message indices after which to apply $\operatorname{repos}$.
Message indices are zero-based, and $\operatorname{drop}$ precedes $\operatorname{repos}$ when both occur at the same message boundary.
For multi-round agent workloads, clients extend the message and cache operation histories after each round, following the standard LLM serving workflow.

\subsection{Cache State Management}
\label{sec:cache-management}

\subsubsection{Design Objectives and Challenges}
\label{sec:design-principles}

\name aims to preserve three key properties of conventional LLM serving systems:
(1) reusing cached states if and only if they are equivalent to the required states,
(2) evicting unused cache states without compromising request isolation,
and (3) restoring evicted cache states within one prefill pass.

Serving systems face three corresponding challenges in preserving these properties when supporting cache engineering:
(1) Token prefixes alone no longer determine cache states, so ordinary prefix matching can incorrectly reuse states from different cache operation histories (Figure~\ref{fig:drop}(b)).
(2) KV cache dropped by one request may still be needed by another, so evicting it without accounting for other requests can cause unnecessary recomputation or violate request isolation.
(3) Ordinary causal prefill cannot reproduce cache states resulting from drops and repositions, while replaying the operation history requires multiple forward passes.
\name addresses these three challenges with an extended radix tree, drop-aware eviction, and segment-batch attention, respectively.

\subsubsection{Drop-aware Cache Management}
\label{sec:serving/drop}

\xhdr{Reuse with drop nodes}
\name extends radix-tree prefix matching with drop nodes.
When a request invokes $\operatorname{drop}$, \name inserts a drop node $\Delta$ at the trigger point along the request's path, recording the logical token indices whose cache is dropped.
Because $\operatorname{drop}$ leaves retained states unchanged but affects subsequent attention computation, this placement preserves prefix sharing while distinguishing later states.
Prefix matching on the extended radix tree reuses cache along matched token and operation prefixes.
In Figure~\ref{fig:drop}(c), trajectories \circlednum{1} and \circlednum{2} share the $A$--$B$ prefix but keep distinct $C$ states because they drop $A$ at different points, while trajectory \circlednum{3} follows a separate branch.
When processing $D$, each trajectory directly reuses its corresponding $B$ and $C$ caches.

\xhdr{Evict with drop-awareness}
\name uses drop information to reclaim unused cache before requests finish.
A cache entry becomes evictable only when no ongoing computation or restoration needs it.
For example, in Figure~\ref{fig:drop}, $A$'s cache can be freed once both requests \circlednum{1} and \circlednum{2} drop it, even as they continue generating $D$.
Eviction frees the physical storage but leaves the tree node as an empty placeholder, keeping cached descendants accessible through prefix matching.
Among evictable cache entries, \name prioritizes evicting dropped cache over cache from completed requests, since the latter is more likely to be reused in the next round.

\xhdr{Restore with segment-batch attention}
\name uses \emph{segment-batch attention} to restore missing cache states in a single prefill pass.
At each layer, queries are split at drop trigger points, and each segment attends causally to its own tokens and to active cache from earlier segments.
Note that dropped cache is excluded from later segments but remains available to earlier segments that originally used it, as shown in Figure~\ref{fig:attention-mask}.
Only attention is segmented, while the FFN processes the full sequence as one batch.
This avoids repeating FFN weight reads, a major source of memory traffic, for each segment.

\subsubsection{Reposition-aware Cache Management}
\label{sec:serving/repos}

While $\operatorname{drop}$ leaves retained cache unchanged, $\operatorname{repos}$ efficiently modifies its keys' position encodings.
Different requests may therefore need the same cache with different position encodings.
Restoration must also reproduce the encodings used at each stage of computation.

\begin{figure}[!tbp]
    \centering
    \includegraphics[width=\linewidth]{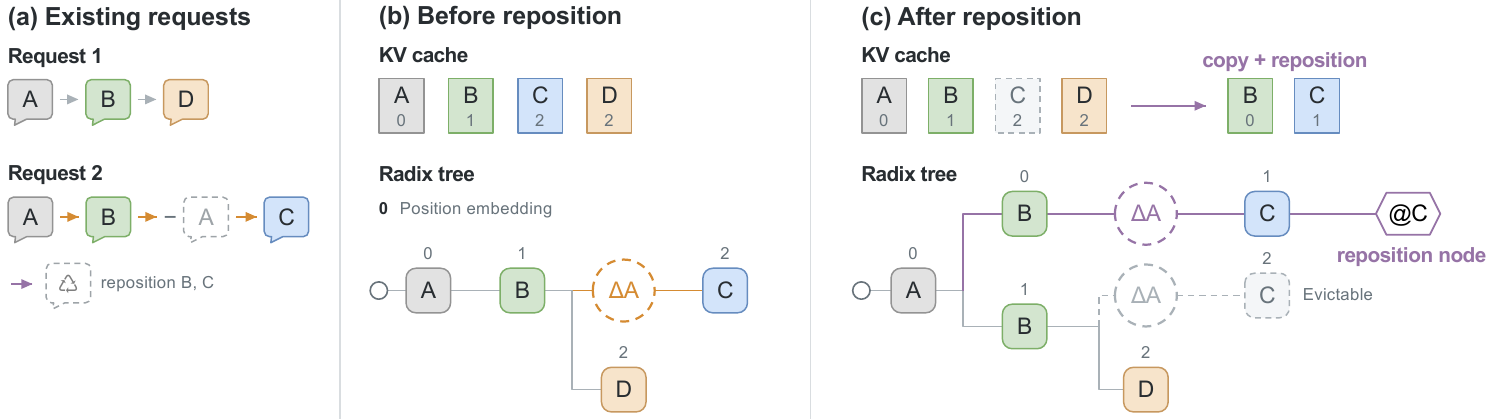}
    \caption{$\operatorname{repos}$ in the KV pool and radix tree.
    (a) Two requests share A and B; Request 2 drops A before appending C.
    (b) B and C initially occupy positions 1 and 2.
    (c) $\operatorname{repos}$ creates copies at positions 0 and 1 by re-encoding K and copying V.
    The original B remains shared with Request 1, while the original C becomes evictable.
    The tree branches at B to distinguish positional versions and records a reposition node after C.
    Labels $p$ denote position $p$.}
    \label{fig:repos}
\end{figure}

\xhdr{Reuse with key rotation}
\name reuses cache after $\operatorname{repos}$ by copying cached states and rotating keys, without recomputing hidden states.
In Figure~\ref{fig:repos}(c), request 2 copies $B$ and $C$ from positions 1 and 2 to positions 0 and 1, while request 1 continues using the original $B$.
In the radix tree, these copies form a new branch with updated position indices and a reposition node after $C$.
Prefix matching checks both token and operation histories: matching position indices allow direct reuse, while differing indices require key rotation.
Appendix~\ref{sec:appendix/system/reposition} explains why reposition nodes are necessary and details the matching rules and position transformations.

\xhdr{Evict positional versions independently}
\name evicts each position encoding versions independently, when it is no longer in use by any ongoing computation or restoration.
For example, in Figure~\ref{fig:repos}(c), the original $B$ remains in use by Request 1, while the original $C$ becomes evictable.

\xhdr{Restore with temporary KV versions}
\name extends segment-batch attention so each segment uses the position encodings from its original computation.
Queries are also split at $\operatorname{repos}$ trigger points.
At each layer, \name rotates keys and copies values to create temporary KV versions for segments that require different positions, which we detail in Appendix~\ref{sec:appendix/system/restore}.

\section{Experiment}
\label{sec:experiment}

\begin{figure}[t]
    \centering
    \begin{minipage}[t]{0.32\linewidth}
        \centering
        \includegraphics[width=\linewidth]{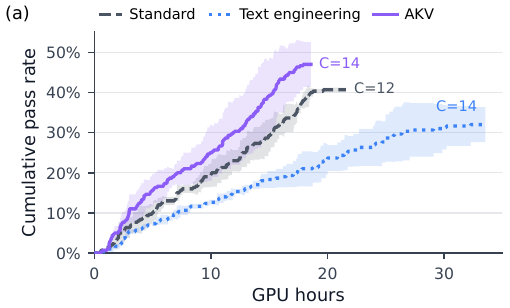}
    \end{minipage}\hfill
    \begin{minipage}[t]{0.32\linewidth}
        \centering
        \includegraphics[width=\linewidth]{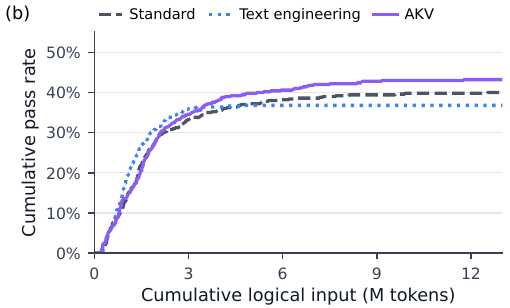}
    \end{minipage}\hfill
    \begin{minipage}[t]{0.32\linewidth}
        \centering
        \includegraphics[width=\linewidth]{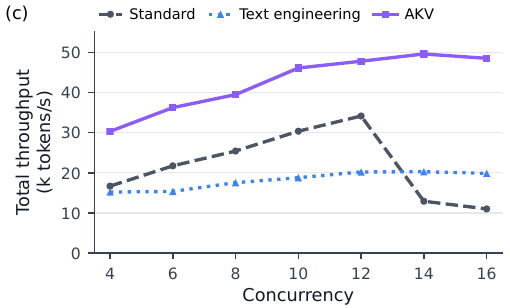}
    \end{minipage}
    \caption{Task performance and serving throughput with GPT-OSS-120B on SWE-bench Verified.
    (a) Online evaluation on the first 100 tasks: mean cumulative pass rate versus GPU hours across three runs per method, with shaded regions indicating $\pm$1 standard deviation.
    (b) Cumulative pass rate versus cumulative logical input tokens over all 500 tasks.
    (c) Serving throughput versus concurrency.}
    \label{fig:performance-overview}
\end{figure}

\subsection{Setup}
\label{sec:experiment/setup}

\xhdr{Baselines}
We compare \name with two context engineering baselines that summarize accumulated history, following common practice in agent harnesses~\citep{anthropic_claudecode_caching,openai_codex_compaction}.
\textit{Standard} uses a higher, model-specific summarization threshold: 96K tokens for GPT-OSS-120B and MiniMax-M2.7, and 192K tokens for MiroThinker-1.7-Mini.
\textit{Text Engineering} uses the same summarization policy with a lower threshold of 48K tokens across all models.
For \name, we use a tool-response rolling-drop policy that retains the KV states of the latest 12 tool responses.
Its $\operatorname{repos}$ threshold matches Standard's summarization threshold for each model.

\xhdr{Models and benchmarks}
We evaluate \name across MiniMax-M2.7~\citep{chen2026minimax}, GPT-OSS-120B~\citep{agarwal2025gpt}, and MiroThinker-1.7-Mini~\citep{team2026mirothinker} LLMs.
GPT-OSS-120B is evaluated with high reasoning effort, while MiniMax-M2.7 and MiroThinker-1.7-Mini are evaluated in thinking mode.
The benchmarks cover various long-horizon agentic tasks, including BrowseComp-Plus~\citep{chen2025browsecompplus} (BCP), xbench-DeepSearch-2510~\citep{xbench2025deepsearch2510} (xbench-DS), and the text-only 103-task
validation subset of GAIA~\citep{mialon2024gaia} (GAIA-Text) for deep research, and SWE-Verified~\citep{openai2024swebenchverified} for software engineering.
Benchmark setup details are provided in Appendix~\ref{sec:appendix/benchmarks}.

\xhdr{Metrics}
We report success rate and mean per-task API cost for the main agent and its context-summary calls, in US cents unless otherwise specified.
Costs are computed from active input and generated output tokens using Groq's rates~\citep{groq_models_pricing} for GPT-OSS-120B and MiniMax's official rates for MiniMax-M2.7.
For MiroThinker-1.7-Mini, we use the official rates of its base model, Qwen3-30B-A3B~\citep{yang2025qwen3}. Detailed token prices are listed in Table~\ref{tab:api-cost}.
We additionally charge each token $\operatorname{repos}$ by \name at the cache-hit rate, deliberately overestimating the cost of key rotation for a conservative comparison.
For serving efficiency, we report request latency, time to first token (TTFT), and time per output token (TPOT), together with input and output throughput.

\xhdr{Serving configuration}
Baselines run on SGLang~\citep{zheng2024sglang}, which is also the base serving system of \name.
Serving experiments use eight NVIDIA A800-80GB GPUs: four for prefill and four for decode, with tensor parallelism of four.
Since A800 GPU lacks native MXFP4 support, all LLM weights are set to BF16.
Further details are in Appendix~\ref{sec:appendix/system/configuration}.

\subsection{Performance}

\begin{table}[t]
  \centering
  \caption{Benchmark scores (pass rate, \%) and costs (cents) by model and management method. Bold indicates the highest score and lowest cost within each model, including ties.}
  \label{tab:experiment-results}
  \setlength{\tabcolsep}{2.5pt}
  \begin{tabularx}{\linewidth}{@{}l*{10}{>{\centering\arraybackslash}X}@{}}
    \toprule
    \multirow{2}{*}{Method}
      & \multicolumn{2}{c}{BCP}
      & \multicolumn{2}{c}{GAIA-Text}
      & \multicolumn{2}{c}{SWE-Verified}
      & \multicolumn{2}{c}{xbench-DS}
      & \multicolumn{2}{c}{Average} \\
    \cmidrule(lr){2-3}\cmidrule(lr){4-5}\cmidrule(lr){6-7}\cmidrule(lr){8-9}\cmidrule(lr){10-11}
      & Score & Cost & Score & Cost & Score & Cost & Score & Cost & Score & Cost \\
    \midrule
    \multicolumn{11}{c}{\textit{MiniMax-M2.7} (230B total, 10B active)} \\
    \midrule
    Standard & 58.9 & 5.2 & 35.0 & \textbf{2.4} & \textbf{75.2} & 7.0 & \textbf{40.5} & 5.5 & 52.4 & 5.0 \\
    Text Engineering & \textbf{62.3} & 5.3 & 36.9 & 2.7 & 72.0 & 9.4 & 40.0 & 5.2 & 52.8 & 5.7 \\
    \rowcolor{gray!10}
    \name & 58.9 & \textbf{4.8} & \textbf{40.8} & 2.6 & 74.4 & \textbf{5.5} & \textbf{40.5} & \textbf{4.7} & \textbf{53.7} & \textbf{4.4} \\

    \midrule
    \multicolumn{11}{c}{\textit{GPT-OSS-120B} (5.1B active)} \\
    \midrule
    Standard & 57.5 & 13.1 & 68.0 & 5.4 & 40.0 & 20.0 & \textbf{38.5} & 6.2 & 51.0 & 11.2 \\
    Text Engineering & 52.7 & 12.6 & 65.1 & 6.8 & 36.8 & 15.8 & \textbf{38.5} & 5.4 & 48.3 & 10.2 \\
    \rowcolor{gray!10}
    \name& \textbf{61.9} & \textbf{11.4} & \textbf{69.9} & \textbf{4.1} & \textbf{43.2} & \textbf{9.7} & 37.5 & \textbf{3.8} & \textbf{53.1} & \textbf{7.3} \\
    \midrule
    \multicolumn{11}{c}{\textit{MiroThinker-1.7-Mini} (30B total, 3B active)} \\
    \midrule
    Standard & 60.0 & 25.2 & 68.0 & \textbf{9.2} & \textbf{7.8} & 33.6 & 40.5 & 13.5 & 44.1 & 20.4 \\
    Text Engineering & 45.7 & 17.6 & 66.0 & 11.0 & 4.2 & 28.4 & 32.5 & 15.3 & 37.1 & 18.1 \\
    \rowcolor{gray!10}
    \name& \textbf{63.5} & \textbf{15.3} & \textbf{68.9} & 9.5 & 7.4 & \textbf{26.3} & \textbf{43.5} & \textbf{8.1} & \textbf{45.8} & \textbf{14.8} \\
    \bottomrule
  \end{tabularx}
\end{table}

\xhdr{Agentic benchmark results}
As shown in Table~\ref{tab:experiment-results}, \name achieves the highest average success rates across three models, improving the average success rate by 1.3--2.1\% and 0.9--8.7\% compared with two baselines, respectively.
\name also reduces inference cost by 27.7\% relative to Standard and 21.9\% relative to Text Engineering, averaged across all three models and four benchmarks.
Appendix~\ref{sec:appendix/cost-analysis} details the saving breakdown.

\xhdr{Source of accuracy improvements}
\name better maintains accuracy beyond the baselines' summarization thresholds.
On the first 100 BCP tasks with GPT-OSS-120B, all three methods achieve comparable accuracy below 48K logical tokens (Figure~\ref{fig:bcp-correct-cost-by-length}).
Text Engineering has the lowest accuracy in the 48--96K range, while Standard is lowest above 96K, following their respective 48K and 96K summarization thresholds.
This pattern is consistent with information loss from abrupt context reductions, whereas rolling drop removes older tool-response cache incrementally, as in Figure~\ref{fig:bcp-input-length-by-round}.

\subsection{Efficiency}

\begin{table}[t]
  \centering
  \caption{Serving performance (latency, s; TTFT, s; TPOT, ms/token; throughput, tokens/s) of GPT-OSS-120B on Top-80 SWE-Verified across concurrency ($C$) levels.
  }
  \label{tab:serving-drop-results}
  \normalsize
  \setlength{\tabcolsep}{2.5pt}
  \renewcommand{\arraystretch}{1.0}
  \begin{tabularx}{\linewidth}{@{}cl*{6}{>{\centering\arraybackslash}X}cc@{}}
    \toprule
    \multirow{2}{*}{$C$}
      & \multirow{2}{*}{Method}
      & \multicolumn{2}{c}{\shortstack{Request latency}}
      & \multicolumn{2}{c}{\shortstack{TTFT}}
      & \multicolumn{2}{c}{\shortstack{TPOT}}
      & \multicolumn{2}{c}{\shortstack{Throughput}} \\
    \cmidrule(lr){3-4}\cmidrule(lr){5-6}\cmidrule(lr){7-8}\cmidrule(lr){9-10}
      & & mean & P95 & mean & P95 & mean & P95 & Input & Output \\
    \midrule
    \multirow{3}{*}{8}
      & Standard & 12.16 & 43.57 & 3.18 & 8.42 & 26.57 & 31.38 & 25200 & 217 \\
      & Text Engineering & 12.84 & 43.57 & 4.94 & 22.99 & \textbf{23.23} & \textbf{25.68} & 17313 & 228 \\
    \rowcolor{gray!10}
      \cellcolor{white} & \name & \textbf{7.79} & \textbf{25.93} & \textbf{1.87} & \textbf{5.99} & 23.50 & 29.42 & \textbf{39227} & \textbf{245} \\
    \midrule
    \multirow{3}{*}{10}
      & Standard & 12.82 & 45.03 & 3.93 & 9.76 & 26.47 & 30.83 & 30108 & 254 \\
      & Text Engineering & 15.17 & 50.27 & 7.35 & 34.07 & \textbf{22.57} & \textbf{25.09} & 18533 & 246 \\
    \rowcolor{gray!10}
      \cellcolor{white} & \name & \textbf{8.13} & \textbf{27.11} & \textbf{2.22} & \textbf{7.36} & 23.20 & 29.86 & \textbf{46002} & \textbf{295} \\
    \midrule
    \multirow{3}{*}{12}
      & Standard & 12.39 & 43.16 & 2.97 & 9.99 & 28.00 & 32.78 & 33832 & 313 \\
      & Text Engineering & 16.83 & 52.82 & 8.86 & 37.98 & 23.24 & \textbf{25.95} & 19945 & 264 \\
    \rowcolor{gray!10}
      \cellcolor{white} & \name & \textbf{8.47} & \textbf{27.18} & \textbf{2.60} & \textbf{8.17} & \textbf{23.06} & 29.23 & \textbf{48923} & \textbf{339} \\
    \midrule
    \multirow{3}{*}{14}
      & Standard & 30.16 & 108.86 & 21.36 & 88.29 & \textbf{24.50} & 29.85 & 12767 & 146 \\
      & Text Engineering & 19.59 & 60.67 & 11.19 & 47.36 & \textbf{24.50} & \textbf{27.23} & 19992 & 267 \\
    \rowcolor{gray!10}
      \cellcolor{white} & \name & \textbf{9.09} & \textbf{28.49} & \textbf{2.92} & \textbf{8.62} & \textbf{24.50} & 31.56 & \textbf{51837} & \textbf{369} \\
    \midrule
    \multirow{3}{*}{16}
      & Standard & 36.50 & 229.90 & 27.71 & 223.57 & 24.83 & 29.25 & 10866 & 138 \\
      & Text Engineering & 23.00 & 69.96 & 14.48 & 52.71 & \textbf{24.44} & \textbf{27.56} & 19595 & 267 \\
    \rowcolor{gray!10}
      \cellcolor{white} & \name & \textbf{10.32} & \textbf{30.63} & \textbf{4.16} & \textbf{13.02} & 24.84 & 32.41 & \textbf{50817} & \textbf{369} \\
    \bottomrule
  \end{tabularx}
\end{table}

\xhdr{Serving efficiency}
As shown in Table~\ref{tab:serving-drop-results}, \name achieves higher throughput and lower request latency than both baselines across concurrency levels.
At their respective peak-throughput settings, \name ($C=14$) achieves $1.36\times$ and $1.51\times$ speedups in mean and P95 request latency over Standard ($C=12$), respectively.
Input and output throughput are $1.53\times$ and $1.18\times$ those of Standard.
Here, input throughput counts all logical input tokens, including those whose cache has been dropped.
The workload and measurement protocol are detailed in Appendix~\ref{sec:appendix/system/evaluation}.
\name's throughput and latency advantages over Standard widen at high concurrency ($C\ge 14$), as KV-cache pressure increases Standard's re-prefill work. As a result, fewer Standard requests can decode concurrently, yielding higher end-to-end latency and lower throughput.

\xhdr{Source of efficiency improvements}
Figure~\ref{fig:decode-breakdown} in Appendix~\ref{sec:appendix/speedup-breakdown} shows that the decode speedup comes mainly from reduced full-attention time.
Dropping older tool-response cache shortens the active context, reducing both KV reads and attention computation at each decode step.

\xhdr{Real-world task acceleration}
In Figure~\ref{fig:performance-overview}(a), \name reaches the highest final pass rate in the least GPU time.
This online evaluation uses GPT-OSS-120B on the first 100 SWE-bench Verified tasks, with each method running at its peak-throughput concurrency.
This advantage combines gains in task performance and serving efficiency.
Figure~\ref{fig:performance-overview}(b) shows that \name continues to improve its pass rate as cumulative logical input tokens increase, ultimately outperforming both baselines.
Figure~\ref{fig:performance-overview}(c) shows higher peak serving throughput, allowing \name to process these trajectories faster.


\subsection{Ablations and Cache Engineering Policies}
\label{sec:experiment/agentic}

\begin{table}[t]
    \centering
    \begin{minipage}[t]{0.52\linewidth}
        \vspace{0pt}
        \caption{Drop-target and $\operatorname{repos}$ ablations with GPT-OSS-120B on each benchmark's first 100 tasks.}
        \label{tab:cache-reduction-ablation}
        \centering
        \normalsize
        \setlength{\tabcolsep}{0.5pt}
        \renewcommand{\arraystretch}{1.0}
        \begin{tabularx}{\linewidth}{@{}lc*{4}{>{\centering\arraybackslash}X}@{}}
            \toprule
            \multirow{2}{*}{Strategy} & \multirow{2}{*}{$\operatorname{Repos}$}
                & \multicolumn{2}{c}{BCP} & \multicolumn{2}{c}{SWE} \\
            \cmidrule(lr){3-4}\cmidrule(lr){5-6}
                & & Score & Cost & Score & Cost \\
            \midrule
            Standard & \xmark & 56.0 & 13.5 & \textbf{41.0} & 19.1 \\
            Tool drop & \xmark & 55.0 & 7.1 & \textbf{41.0} & 9.4 \\
            Reasoning drop & \cmark & 56.5 & 8.2 & 36.5 & 14.2 \\
            Both drop & \cmark & 56.5 & 8.7 & 34.5 & 7.8 \\
            \rowcolor{gray!10}
            Tool drop & \cmark & \textbf{62.0} & 10.9 & \textbf{41.0} & 9.5 \\
            \bottomrule
        \end{tabularx}
    \end{minipage}\hfill
    \begin{minipage}[t]{0.46\linewidth}
        \vspace{0pt}
        \caption{Agentic cache reduction on a 100-task BCP subset. CAT means Cache-As-Tool.}
        \label{tab:context-management-tool}
        \centering
        \normalsize
        \setlength{\tabcolsep}{0.5pt}
        \renewcommand{\arraystretch}{1.097}
        \begin{tabular*}{\linewidth}{@{\extracolsep{\fill}}clccc@{}}
            \toprule
            \multirow{2}{*}{Model} & \multirow{2}{*}{Metric}
                & \multicolumn{3}{c}{Method} \\
            \cmidrule(lr){3-5}
                & & Standard & Text Eng. & \cellcolor{gray!10}CAT \\
            \midrule
            \multirow{2}{*}{\shortstack[c]{GPT-OSS-\\120B}}
                & Score & 52.1 & 50.0 & \cellcolor{gray!10}\textbf{61.0} \\
                & Cost & 13.4 & 13.9 & \cellcolor{gray!10}\textbf{12.1} \\
            \midrule
            \multirow{2}{*}{\shortstack[c]{MiniMax-\\M2.7}}
                & Score & 56.0 & 61.0 & \cellcolor{gray!10}\textbf{64.0} \\
                & Cost & \textbf{5.4} & 5.5 & \cellcolor{gray!10}5.9 \\
            \bottomrule
        \end{tabular*}
    \end{minipage}
\end{table}

\xhdr{Ablation study}
Table~\ref{tab:cache-reduction-ablation} compares drop targets and the use of $\operatorname{repos}$, showing that tool-response drop with $\operatorname{repos}$ achieves the highest pass rate on both benchmarks.
Adding $\operatorname{repos}$ raises BCP's pass rate from 55\% to 62\%, allowing exploration beyond the position limit that would otherwise halt execution.
With $\operatorname{repos}$ enabled, dropping reasoning cache, either alone or together with tool-response cache, yields lower pass rates on both benchmarks than dropping tool-response cache alone.

\xhdr{Recursive self-improvement (RSI)}
Figure~\ref{fig:minimax-self-evolution} shows that RSI finds lower-cost cache policies than our manually designed tool-response rolling-drop policy ($K=12$) while maintaining Standard's accuracy.
We use MiniMax-M2.7 to refine policies on 20 BCP tasks and evaluate the resulting policies on 100 held-out tasks.
The policies from iterations 3 and 6 reduce per-task cost by 15.6\% and 17.8\% relative to Standard, respectively.

\xhdr{Cache management as a tool}
Exposing cache management as a tool lets agents proactively decide when to reduce the cache and what to retain.
Table~\ref{tab:context-management-tool} shows that this approach achieves the highest pass rate on a 100-task BCP subset for both models: 61.0\% for GPT-OSS-120B and 64.0\% for MiniMax-M2.7.
Compared with Standard, GPT-OSS-120B's per-task cost falls from 13.4 to 12.1 cents, while MiniMax-M2.7's rises from 5.4 to 5.9 cents as it executes more turns.

\section{Conclusion}

We introduce cache engineering, enabling agents to directly manage active KV states while reusing retained caches without recomputation.
We co-design the \name serving system to support cache engineering through a stateless interface, an extended radix tree, and segment-batch attention for efficient cache reuse, eviction, and restoration.
Across agent benchmarks, cache engineering improves average task performance while reducing inference costs, with \name translating these efficiency gains into higher serving throughput and lower latency.

\xhdr{Limitations}
Further exploration of cache engineering strategies, training agents under this scheme, and extending serving support to linear attention models remain future work.

\clearpage

\section*{Acknowledgements}

This work was supported by National Natural Science Foundation of China (No. 62506197, 62325405, 62104128, U19B2019, U21B2031, 61832007, 62204164, 92364201), Tsinghua EE Xilinx AI Research Fund, and Beijing National Research Center for Information Science and Technology (BNRist).
We thank all the support from Infinigence-AI.




\bibliography{main}
\bibliographystyle{iclr2027_conference}

\clearpage
\appendix
\section{Appendix}
\label{sec:appendix}

\subsection{\name System Design Details}
\label{sec:appendix/system}

\subsubsection{Restore with Segment-batch Attention}
\label{sec:appendix/system/restore}

\xhdr{Segmentation}
Building on Sections~\ref{sec:serving/drop} and~\ref{sec:serving/repos}, \name restores missing KV states by partitioning queries whenever a $\operatorname{drop}$ or $\operatorname{repos}$ operation changes the visible history or its position mapping.
Each segment selects the corresponding KV pages and applies standard causal attention, encoding historical visibility without constructing a dense custom mask.

\begin{figure}[!htbp]
    \centering
    \includegraphics[width=0.7\linewidth]{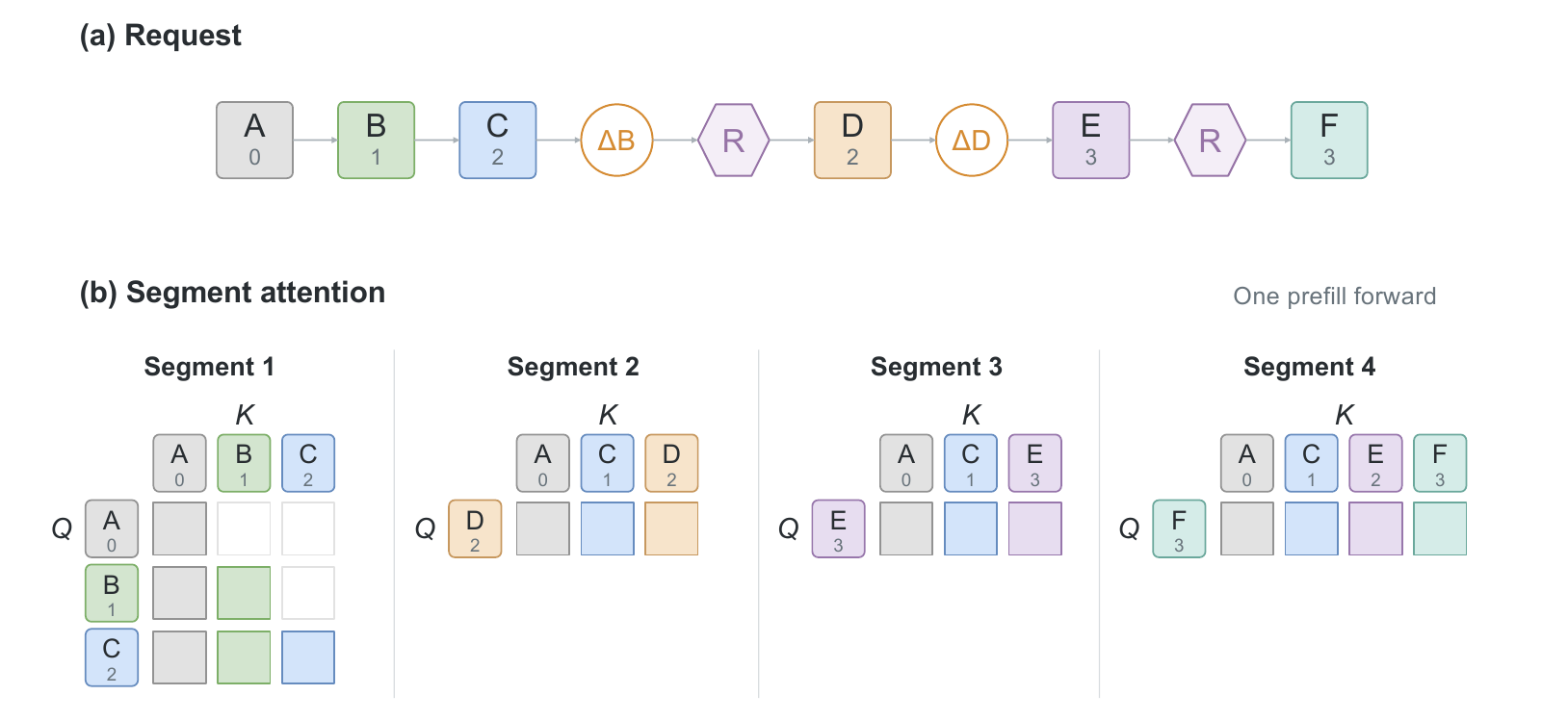}
    \caption{\textbf{Restore with segment-batch attention.}
    The request in (a) produces the four attention segments in (b).
    Numbers denote query/key positions; colored cells indicate allowed attention.
    }
    \label{fig:attention-mask}
\end{figure}

\xhdr{Example}
Figure~\ref{fig:attention-mask} illustrates a complete restoration, treating each letter as one token and writing $X_p$ for token $X$ at position $p$.
Segment 1 computes $A_0,B_1,C_2$ with causal attention.
After dropping $B$ and repositioning, Segment 2 computes $D_2$ using keys $(A_0,C_1,D_2)$.
Dropping $D$ starts Segment 3: $E_3$ attends to $(A_0,C_1,E_3)$.
Although Segments 2 and 3 share a position mapping, they require separate KV views because $D$ is no longer visible to $E$; $\operatorname{drop}$ alone does not compact positions.
The second $\operatorname{repos}$ moves $E$ from position 3 to 2, so Segment 4 computes $F_3$ using $(A_0,C_1,E_2,F_3)$.

\xhdr{Batched restoration}
Within each layer, \name writes the newly computed KV states and constructs the additional position versions, such as $C_1$ and $E_2$, by re-encoding K and copying V.
The segment-specific page lists and query ranges are then submitted together to batched attention.
Thus, all four segments execute within a single model forward pass, without recomputing hidden states for each position version or replaying separate forward passes at operation boundaries.

\subsubsection{Reposition Scheme}
\label{sec:appendix/system/reposition}

\xhdr{Design objective}
Reposition-aware cache management must distinguish positional versions while reusing their shared computational states.
We therefore first identify directly reusable caches via exact matching, then use retrigger matching to find prefixes that are reusable via positional transformation, avoiding unnecessary hidden-state recomputation.

\begin{figure}[t]
    \centering
    \includegraphics[width=\linewidth]{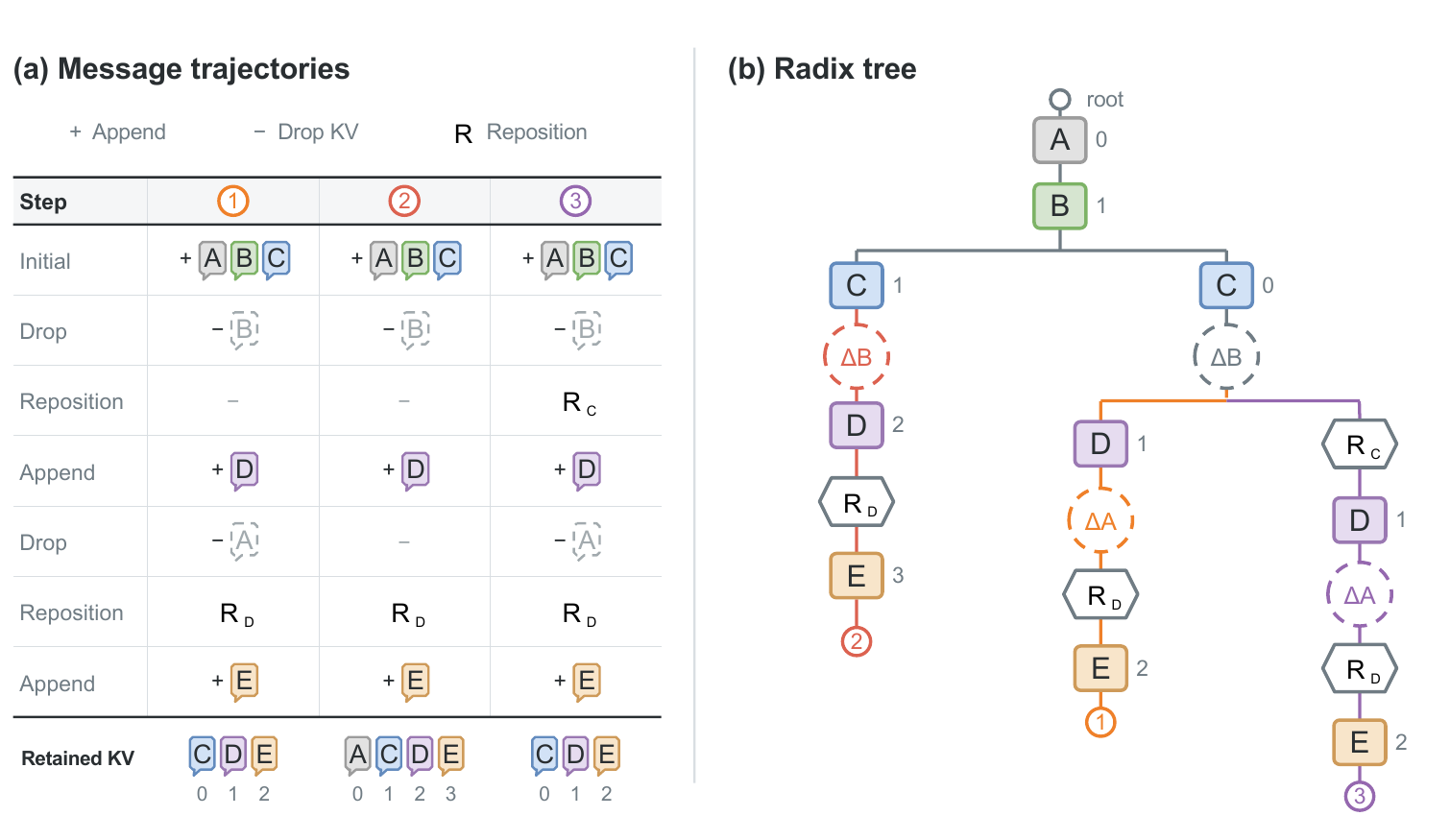}
    \caption{\textbf{Encoding drop and reposition histories in the extended radix tree.}
    (a) Three message trajectories interleave token appends, KV drops, and reposition operations.
    Although their retained tokens overlap, applying the operations at different points produces different KV states.
    (b) \name records each operation in the radix tree, distinguishing cache versions by position and operation history.
    Numbers beside token nodes denote stored position encoding indices; $R_X$ denotes $\operatorname{repos}$ after $X$.
    }
    \label{fig:reposition-radix-tree}
\end{figure}

\xhdr{Token-node representation}
Each token node is represented as $T(x,p)$, where $x$ is the token ID and $p$ is the position currently encoded in its cached K.
The position label describes the stored cache, not necessarily the position at which the token was originally computed.
For example, $C$ is initially computed at position 2 in all three trajectories in Figure~\ref{fig:reposition-radix-tree}.
After $\operatorname{repos}$, trajectories 1 and 3 use $(C,0)$, while trajectory 2 uses $(C,1)$.
Operation nodes along the path preserve the operation history that governs computation.

\xhdr{Why position labels and reposition nodes are needed}

Position labels distinguish cache representations that cannot be used interchangeably.
For example, $C$ has the same original computation history in trajectories 1 and 2, but is repositioned to 0 and 1, respectively.
Its K encodings therefore differ and require positional transformation before reuse.

Reposition nodes distinguish states that share final positions but arise from different computations.
Trajectories 1 and 3 both drop $B$ after $ABC$, but trajectory 3 executes $R_C$ before computing $D$.
When $D$ is computed, the positions of $A$, $C$, and $D$ are respectively $(0,2,3)$ and $(0,1,2)$.
The relative distance between $D$ and $A$ changes, while that between $D$ and $C$ remains unchanged, potentially producing different hidden states and KV entries for $D$.
Both trajectories subsequently drop $A$ and execute $R_D$ to move $C$ and $D$ to positions 0 and 1.
This later transformation cannot erase the computational difference already established in $D$.
The path must therefore retain $R_C$ before $D$ to prevent incorrect reuse of $D$ and subsequent cached states.

\xhdr{Exact and retrigger matching}
Exact matching compares all token-node fields:
\[
T(x,p)\equiv T(x',p')
\iff x=x' \land p=p',
\]
and requires identical operation nodes and their parameters.
Retrigger matching ignores only stored position encoding indices while preserving all operation records:
\[
\phi(T(x,p))=T(x),\qquad
\phi(\Delta P)=\Delta P,\qquad
\phi(R)=R.
\]
Two prefixes are history-compatible if their projections are identical.

The system initializes the result with the longest valid exact prefix, then searches compatible branches from the root.
A candidate replaces the current result only if it is valid and matches a strictly longer prefix, preserving exact matches in a tie.
Retrigger matching stops at the first token or operation mismatch.
Selected source caches are copied and positionally transformed as needed before use.

\xhdr{Correctness of retrigger matching}
Under fixed model and position encoding configurations and exact arithmetic, identical projected prefixes preserve the token and operation history preceding each token's original computation.
They therefore produce the same hidden states and unrotated KV content.
$\operatorname{Repos}$ operations applied afterward change only K's positional representation, allowing a source version $s$ to be transformed into a target version $t$ as
\[
K_t=R(p_t)R(p_s)^{-1}K_s,
\qquad V_t=V_s,
\]
without recomputing hidden states.

Preserving operation nodes prevents this transformation from crossing a divergence in computational history.
For example, trajectories 1 and 3 disagree at $R_C$ before $D$, so retrigger matching cannot reuse $D$ across them.
Finite-precision transformations may introduce rounding error; this correctness argument does not require bitwise equivalence to recomputation.

\subsubsection{Serving Configuration}
\label{sec:appendix/system/configuration}

\name is implemented on SGLang v0.5.20.
Table~\ref{tab:appendix-serving-configuration} summarizes the configuration used for the serving experiments.
All three methods use GPT-OSS-120B with weights expanded to BF16 on eight A800 GPUs, with separate prefill and decode workers.
Both workers enable radix caching; the baselines use LRU eviction, while \name enables drop-aware eviction.

\begin{table}[htbp]
    \centering
    \caption{\textbf{Serving configuration.} }
    \label{tab:appendix-serving-configuration}
    \small
    \renewcommand{\arraystretch}{1.12}
    \begin{tabularx}{\linewidth}{@{}lX@{}}
        \toprule
        Setting & Value \\
        \midrule
        Model and weight precision & GPT-OSS-120B, BF16 \\
        Hardware & Eight NVIDIA A800 GPUs, 80\,GB each \\
        Parallelism & Prefill: TP4; decode: TP4 \\
        KV transfer & Mooncake over intra-node NVLink \\
        Attention backend & Triton \\
        MoE backend & \texttt{triton\_kernel} \\
        Static memory fraction & 0.85 \\
        KV page size & One token per page \\
        Prefill chunk size & 8,192 tokens \\
        Maximum prefill tokens & 8,192 \\
        Prefix caching & Radix cache enabled on both workers \\
        Eviction & Baselines: LRU; \name: drop-aware eviction \\
        \bottomrule
    \end{tabularx}
\end{table}

\subsubsection{Workload Replay and Evaluation Protocol}
\label{sec:appendix/system/evaluation}

\xhdr{Task selection}
To emulate realistic long-horizon agent workloads with substantial accumulated context, we select the 80 SWE-bench Verified tasks with the largest context lengths in trajectories, without any summary operation. These tasks reflect the difficulty of tasks in long-context reasoning.
We rank tasks by this peak length and shuffle the selected tasks with a fixed seed.
All methods use the same tasks and ordering.
Each method replays trajectories collected under its own context/cache engineering policy, so the number of turns and the output budgets may differ between methods.

\xhdr{Trajectory replay}
For each turn, we reconstruct the output token budget from the complete recorded message history, including reasoning and tool-call content, after subtracting the generation prefix already present in the input.
We use temperature 0 and \texttt{ignore\_eos=True} to generate the prescribed number of tokens.
The newly generated assistant response is retained in the history for subsequent turns, while tool responses are replayed from the recorded trajectory rather than re-executing the tools.
Recorded terminal inputs without an available assistant response are excluded from replay.

\xhdr{Concurrency and measurement windows}
We maintain $C$ client task concurrency and execute turns sequentially within each task.
The task-count target for each measurement window is $\max(3 \times C, 30)$; measurements include requests from supplemental tasks used to maintain concurrency.
For \name at $C\in\{14,16\}$, we truncate the archived runs at the completion of the first 30 tasks in completion order, counting all requests within the resulting window rather than only requests from those 30 completed tasks.
For Standard at large concurrency, KV cache capacity pressure causes repeated prefill, resulting in severe serving degradation.
We therefore report these two runs over a fixed one-hour window, with some requests still unfinished at the cutoff.

\subsection{Experimental Details}
\label{sec:appendix/experimental-details}

\subsubsection{Evaluation Configurations}
\label{sec:appendix/benchmarks}

\xhdr{Evaluation harnesses}
We adapt the official BrowseComp-Plus harness, mini-SWE-agent~\citep{yang2024sweagent}, the browser-agent scaffold from observation-masking~\citep{zhang2026masking}, and WebExplorer's evaluation harness~\citep{liu2025webexplorer} for BCP, SWE-bench Verified, GAIA-Text, and xbench-DeepSearch-2510, respectively.

\xhdr{Retrieval and webpage reading}
BCP evaluation follows the official task prompts and uses Qwen3-Embedding-8B~\citep{zhang2025qwen3embedding} to retrieve documents from the benchmark's fixed corpus.
All models and methods have access to both \texttt{search} and the official optional \texttt{get\_document} tool, which returns the full text of a selected document.
For xbench-DeepSearch-2510, we use GLM-5.2~\citep{zai2026glm52} to extract information from webpages in response to the agent's queries.

\xhdr{Answer grading}
We use GLM-5.2 as the answer judge for all three question-answering benchmarks.
The grading prompts follow the official BCP implementation for BCP, observation-masking for GAIA-Text, and WebExplorer for xbench-DeepSearch-2510.
For SWE-bench Verified, submitted patches are evaluated using the official test scripts and grading functions; no LLM judge is used.

\xhdr{Agent step limits}
The agent step limits are 100 for BCP, 250 for SWE-bench Verified, 200 for GAIA-Text, and 100 for xbench-DeepSearch-2510.

\subsubsection{Context Engineering Baseline Details}
\label{sec:appendix/summarization}

Our context engineering baselines adopt Pi's summarization prompts, conversation serialization, cut-point selection, and summary-generation budgets~\citep{earendil_works_2026_pi}, with \texttt{reserveTokens}=16,384 and \texttt{keepRecentTokens}=20,000 across experimental groups.
Summarization uses the agent's model and reasoning settings, replacing older dialogue with a summary while preserving system and developer instructions and the selected recent messages.

Unlike Pi's native trigger based on the context window minus the reserve-token budget, we trigger compaction when the exact prompt token count reaches the threshold specified in Section~\ref{sec:experiment/setup}.
We also reject empty summaries and handle summarization failures according to the benchmark error protocol without retry.

Pi defines four prompt templates: a shared system prompt, an initial-summary prompt, an update-summary prompt, and a turn-prefix prompt.
For brevity, we reproduce the shared system and turn-prefix prompts below.
The turn-prefix prompt is appended to the serialized history in the summary request's user message.

\begin{bluebox}[Shared system prompt]{width=\linewidth, valign=top}
\small\ttfamily\raggedright
You are a context summarization assistant. Your task is to read a conversation between a user and an AI assistant, then produce a structured summary following the exact format specified.\par\medskip
Do NOT continue the conversation. Do NOT respond to any questions in the conversation. ONLY output the structured summary.
\end{bluebox}

\begin{bluebox}[Turn-prefix prompt]{width=\linewidth, valign=top}
\small\ttfamily\raggedright
This is the PREFIX of a turn that was too large to keep. The SUFFIX (recent work) is retained.\par\medskip
Summarize the prefix to provide context for the retained suffix:\par\medskip
\#\# Original Request\par
[What did the user ask for in this turn?]\par\medskip
\#\# Early Progress\par
- [Key decisions and work done in the prefix]\par\medskip
\#\# Context for Suffix\par
- [Information needed to understand the retained recent work]\par\medskip
Be concise. Focus on what's needed to understand the kept suffix.
\end{bluebox}

\subsection{API Cost Accounting for Experiment}
\label{sec:appendix/api-cost}

We estimate API-equivalent inference costs using the fixed token prices listed in Table~\ref{tab:api-cost}.
Our cost accounting includes only the evaluated agent's generations and its context-summary calls.
We use Groq's rates~\citep{groq_models_pricing} for GPT-OSS-120B and GPT-OSS-20B, and MiniMax's official rates for MiniMax-M2.7.
For MiroThinker-1.7-Mini, we use the official rates of its base model, Qwen3-30B-A3B~\citep{yang2025qwen3}.
Cache-miss, cache-hit, and output tokens are charged separately at different rates, while using the same prices across different methods for the same model.
For \name's cache repositioning, we also charge it at the cache-hit rate as a conservative estimate of the cost of key rotation.

\begin{table}[htbp]
    \centering
    \caption{Token prices for API-equivalent cost rate (USD per million tokens).}
    \label{tab:api-cost}
    \setlength{\tabcolsep}{2.5pt}
    \begin{tabularx}{\linewidth}{@{}l*{3}{>{\centering\arraybackslash}X}@{}}
        \toprule
        Model & Cache-miss & Cache-hit & Output \\
        \midrule
        MiniMax-M2.7 & 0.300 & 0.060 & 1.200 \\
        GPT-OSS-120B & 0.150 & 0.075 & 0.600 \\
        MiroThinker-1.7-Mini & 0.200 & 0.040 & 2.400 \\
        GPT-OSS-20B & 0.075 & 0.0375 & 0.300 \\
        \bottomrule
    \end{tabularx}
\end{table}

\subsection{Additional Experiment Results}

\subsubsection{Speedup Breakdown}
\label{sec:appendix/speedup-breakdown}
\begin{figure}[htbp]
    \centering
    \begin{minipage}{0.6\linewidth}
        \centering
        \includegraphics[width=\linewidth]{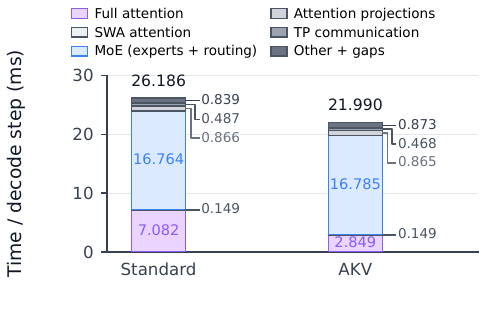}
        \caption{\textbf{Latency breakdown.}
        Decode-step time for GPT-OSS-120B at the same batch size, broken down by operation.}
        \label{fig:decode-breakdown}
    \end{minipage}
\end{figure}

\name accelerates decoding primarily by reducing full-attention time.
As shown in Figure~\ref{fig:decode-breakdown}, at the same batch size, \name reduces full-attention time by 59.8\% and total decode-step time by 16.0\% relative to Standard.
Dropping older tool-response KV shortens the active context, so subsequent decode steps load less KV data from GPU memory and perform fewer attention computations.

\subsubsection{Cost Breakdown and Active Context Length}
\label{sec:appendix/cost-analysis}

\begin{figure}[tb]
    \centering
    \begin{minipage}[t]{0.48\linewidth}
        \vspace{0pt}
        \centering
        \includegraphics[width=\linewidth]{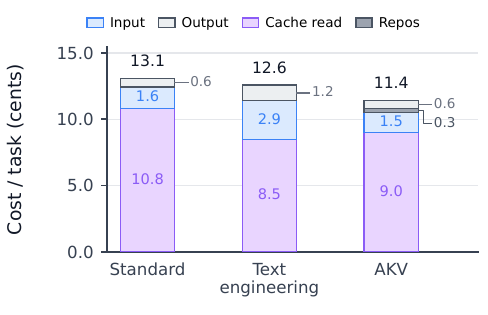}
        \caption{\textbf{Cost breakdown.}
        Average cost per task on BCP with GPT-OSS-120B for Standard, Text Engineering, and \name.}
        \label{fig:bcp-cost-breakdown}
    \end{minipage}\hfill
    \begin{minipage}[t]{0.48\linewidth}
        \vspace{0pt}
        \centering
        \includegraphics[width=\linewidth]{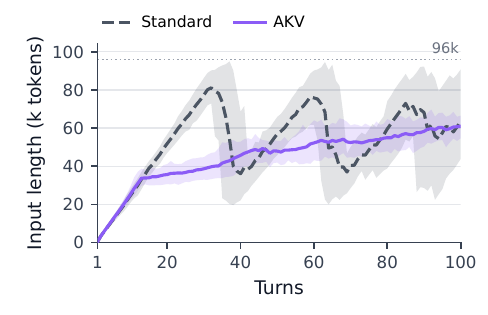}
    \caption{\textbf{Active context length by turn.}
    Standard and \name on BCP over 100 turns. The dotted line marks 96k tokens.}
    \label{fig:bcp-input-length-by-round}
    \end{minipage}
\end{figure}

\xhdr{Cost breakdown}
\name lowers costs by reducing cache reads and repeated prefill.
As shown in Figure~\ref{fig:bcp-cost-breakdown}, dropping older tool-response KV reduces the amount of cached history read by subsequent requests, lowering cache-read costs by approximately 17\% relative to Standard.
Compared with Text Engineering, \name reuses retained KV directly without the re-prefill required when summarization rewrites the history.
This reduces re-prefill computation and lowers cache-miss input costs by approximately 48\%.

\xhdr{Active context length}
Figure~\ref{fig:bcp-input-length-by-round} shows how \name slows the growth of active context.
Once tool responses exceed the retention budget of $K$ rounds, \name continually drops older tool-response KV while retaining the most recent $K$ rounds and reasoning KV.
The active context therefore continues to grow as reasoning accumulates, but at a substantially slower rate.
In contrast, Standard accumulates history until summarization triggers a sharp reduction in context length.

\subsubsection{Attention Across Message History}

\begin{figure}[htbp]
    \centering
    \includegraphics[width=\linewidth]{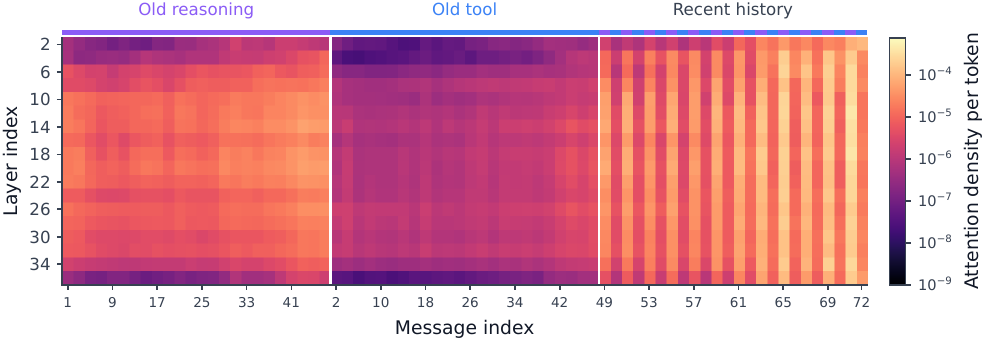}
    \caption{\textbf{Attention density across message history.}
    Attention density per token by layer and message index averaged over 10 selected BrowseComp-Plus trajectories, divided into old reasoning, old tool responses, and recent history.}
    \label{fig:bcp-attention-by-message}
\end{figure}

To explain why our tool-response rolling-drop policy prioritizes dropping older tool-response KV entries, we analyze the attention allocated to different types of historical messages.
Figure~\ref{fig:bcp-attention-by-message} shows attention per token across layers averaged over 10 BrowseComp-Plus trajectories.
Older tool responses receive less attention than older reasoning in most displayed layers, while recent history generally receives more attention.
This pattern suggests that direct attention to historical messages depends on both recency and message type, providing empirical support for retaining recent tool responses and reasoning while preferentially dropping older tool-response KV entries.
Together with the task performance and cost results reported earlier, this observation helps explain how \name reduces KV-cache access while maintaining task performance.

\xhdr{Compatibility with Harness}
To assess whether \name improves an existing harness, we integrate it with Pi~\citep{earendil_works_2026_pi} and evaluate the combination on BCP. For both GPT-OSS-120B and GPT-OSS-20B, adding \name improves accuracy by 3\% over the native Pi harness without compaction and 4\% over the Pi harness with compaction.

\subsubsection{Additional Models}

\begin{table}[h]
  \centering
  \caption{Benchmark scores (pass rate, \%) and costs (cents) of GPT-OSS-20B across different management methods. Bold indicates the highest score and lowest cost, including ties.}
  \label{tab:experiment-results-20b}
  \setlength{\tabcolsep}{2.5pt}
  \begin{tabularx}{\linewidth}{@{}l*{10}{>{\centering\arraybackslash}X}@{}}
    \toprule
    \multirow{2}{*}{Method}
      & \multicolumn{2}{c}{BCP}
      & \multicolumn{2}{c}{GAIA-Text}
      & \multicolumn{2}{c}{SWE-Verified}
      & \multicolumn{2}{c}{xbench-DS}
      & \multicolumn{2}{c}{Average} \\
    \cmidrule(lr){2-3}\cmidrule(lr){4-5}\cmidrule(lr){6-7}\cmidrule(lr){8-9}\cmidrule(lr){10-11}
      & Score & Cost & Score & Cost & Score & Cost & Score & Cost & Score & Cost \\
    \midrule
    Standard & 49.0 & 6.5 & 60.2 & 3.5 & 41.8 & 12.5 & \textbf{25.0} & 2.4 & 44.0 & 6.2 \\
    Text Engineering & 43.0 & \textbf{4.4} & \textbf{65.1} & 3.0 & 25.6 & \textbf{8.8} & 19.0 & 2.0 & 38.2 & \textbf{4.6} \\
    \rowcolor{gray!10}
    \name& \textbf{50.5} & 6.0 & 62.1 & \textbf{2.5} & \textbf{44.2} & 8.9 & 24.0 & \textbf{1.8} & \textbf{45.2} & 4.8 \\
    \bottomrule
  \end{tabularx}
\end{table}

We further evaluate \name on GPT-OSS-20B~\citep{agarwal2025gpt}, a smaller model from the same model family as GPT-OSS-120B, and compare its pass rate and cost against the Standard and Text Engineering baselines.
GPT-OSS-20B is evaluated with high reasoning effort across all methods.
We use the same thresholds as for GPT-OSS-120B: 96K tokens for Standard summarization and \name's $\operatorname{repos}$, and 48K tokens for Text Engineering.
As shown in Table~\ref{tab:experiment-results-20b}, \name achieves the highest average pass rate of 45.2\%, while maintaining a comparable average cost of 4.8 cents. In particular, \name achieves the best pass rate on BCP and SWE-Verified, and the lowest cost on GAIA-Text and xbench-DS. On several metrics where \name does not outperform the baselines, we attribute the smaller gains to the more limited capability of GPT-OSS-20B: compared with GPT-OSS-120B, the smaller model is less able to effectively leverage the retained information under cache engineering, making it more sensitive to cache modifications.

\subsection{Additional Related Works}
\label{sec:concurrent_works}

\xhdr{Concurrent works}
Concurrent work adapts KV eviction to the changing attention needs of agent trajectories.
AgentKV~\citep{liu2026agentkv} scores KV entries using query buffers from different execution phases, retaining information that recency-based scoring can miss.
IntentKV~\citep{li2026intentkv} combines cross-turn query memory with learned importance estimates to preserve useful history under limited KV budgets.
SideQuest~\citep{kariyappa2026sidequest} explores model-directed retention through an auxiliary reasoning branch that identifies stale tool results and issues cache-deletion commands, but its current strategy is limited to tool responses, and its serving integration is tailored to this specific eviction workflow.
Context Language Models (CLMs)~\citep{shao2026clm} primarily study model-driven context management, treating the live context as an editable file and learning how to maintain it.
CLM also introduces Suffix Cache Reuse to preserve unchanged downstream KV states after text edits.
Its serving support centers on session-local reuse: relocated states remain in private buffers outside the shared radix cache, without a unified mechanism for indexing, managing, and recovering these history-dependent states.
\name addresses these serving concerns directly by exposing policy-independent cache operations through a stateless interface and uses an extended radix tree to track both tokens and cache operation histories, integrating retained states into shared cache reuse, memory management, and recovery.

\end{document}